\def\year{2017}\relax
\documentclass[letterpaper]{article}
\usepackage{aaai17}
\usepackage{times}
\usepackage{helvet}
\usepackage{courier}
\usepackage{color}
\usepackage[dvipsnames, table]{xcolor}

\usepackage{tikz}
\usetikzlibrary{fit,automata,positioning,shapes,shapes.multipart,
arrows,decorations.pathreplacing,matrix} 
\usetikzlibrary{decorations.pathmorphing}
\usetikzlibrary{shapes.callouts} 
\usetikzlibrary{decorations.text}
\usetikzlibrary{patterns} 
\tikzset{text/.default=}
\tikzset{node/.default=}
\usetikzlibrary{matrix}
\usepackage{amssymb}
\usepackage{pifont}
\newcommand{\cmark}{\textcolor{ForestGreen}{\ding{51}}}%
\newcommand{\xmark}{\textcolor{Maroon}{\ding{55}}}%

\usepackage{pgfplots}
\usepackage{pgfkeys}
\pgfplotsset{compat=1.3}

\usepackage{listings} 
\usepackage{enumerate}

\usepackage{enumitem}
\newenvironment{my_enumerate}{
   \begin{enumerate}[leftmargin=*]
   \setlength{\itemsep}{0pt}
   \setlength{\parskip}{0pt}
   \setlength{\parsep}{0pt}
}{
   \end{enumerate}
}

\setitemize{noitemsep,topsep=0pt,parsep=0pt,partopsep=0pt, leftmargin=12pt}

\usepackage{array,graphicx}
\usepackage{pifont}
\usepackage{booktabs}
\usepackage{makecell}
\usepackage{csvsimple}
\usepackage{pgfplotstable,wrapfig,lipsum}

\newcommand{\sat}{\texttt{Satisfactory}}
\newcommand{\satshort}{\texttt{Sat.}}
\newcommand{\unsat}{\texttt{Unsatisfactory}}
\newcommand{\unsatshort}{\texttt{Unsat.}}
\newcommand{\eval}{\texttt{Evaluation}}
\newcommand{\evalshort}{\texttt{Eval.}}
\newcommand{\val}{\texttt{Validation}}
\newcommand{\valshort}{\texttt{Val.}}

\newcommand{\Simley}[1]{%
\begin{tikzpicture}[scale=0.15]
    \newcommand*{\SmileyRadius}{1.0}%
    \draw [fill=brown!10] (0,0) circle (\SmileyRadius)
        ;  

    \pgfmathsetmacro{\eyeX}{0.5*\SmileyRadius*cos(30)}
    \pgfmathsetmacro{\eyeY}{0.5*\SmileyRadius*sin(30)}
    \draw [fill=black,draw=none] (\eyeX,\eyeY) circle (0.15cm);
    \draw [fill=black,draw=none] (-\eyeX,\eyeY) circle (0.15cm);

    \pgfmathsetmacro{\xScale}{2*\eyeX/180}
    \pgfmathsetmacro{\yScale}{1.0*\eyeY}
    \draw[color=black, domain=-\eyeX:\eyeX]   
        plot ({\x},{
            -0.1+#1*0.15 
            -#1*1.75*\yScale*(sin((\x+\eyeX)/\xScale))-\eyeY});
\end{tikzpicture}%
}%

\begin{document}
%
\title{On Human Intellect and Machine Failures:\\ Troubleshooting Integrative
Machine Learning Systems}

\author{
Besmira Nushi$^1$ \quad Ece Kamar$^2$ \quad Eric Horvitz$^2$ \quad Donald
Kossmann$^{1,2}$\\
\\
$^1$ETH Zurich, Department of Computer Science, Switzerland\\
$^2$Microsoft Research, Redmond, WA, USA
}
\maketitle
\begin{abstract}
We study the problem of troubleshooting machine learning systems that rely on analytical
pipelines of distinct components. Understanding and fixing errors that arise in
such integrative systems is difficult as failures can occur at multiple
points in the execution workflow. Moreover, errors can propagate, become amplified
or be suppressed, making blame assignment difficult. We propose a
human-in-the-loop methodology which leverages human intellect for
troubleshooting system failures. The
approach simulates potential component fixes through human
computation tasks and measures the
expected improvements in the holistic behavior of the system. The method provides guidance to designers about how they
can best improve the system. We demonstrate the effectiveness of the approach on
 an automated image captioning system that has been pressed into real-world use.
\end{abstract}
\section{Introduction}
Advances in machine learning have enabled the design of integrative systems that
perform sophisticated tasks via the execution of analytical pipelines of
components. Despite the widespread adoption of such systems, current
applications lack the ability to understand, diagnose, and fix their own
mistakes which consequently reduces users' trust and limits future improvements.
Therefore, the problem of understanding and troubleshooting failures of machine
learning systems is of particular interest in the community
\cite{sculley2015hidden}. Our work studies \emph{component-based} machine
learning systems composed of specialized components that are individually
trained for solving specific problems and work altogether for solving a single
complex task. We analyze how the special characteristics of
these integrated learning systems, including \emph{continuous} (non-binary)
success measures, \emph{entangled} component design, and \emph{non-monotonic}
error propagation, make it challenging to assign blame to individual components.
These challenges hinder future system improvements as designers lack an
understanding of how different potential fixes on components may improve the
overall system output.

\begin{figure}[t]
\centering
\begin{tikzpicture}
	\footnotesize
	\tikzstyle{text}=[draw=none,fill=none]
	\tikzstyle{box}=[rectangle, draw=MidnightBlue!100]
	
    \node[box] (system) [
    minimum width = 60 mm, minimum height = 34 mm]{};
    \node[box] (c1) [above left= -1.2 cm and -1.6 cm of system,
    minimum width = 14 mm, minimum height = 10 mm, text width=7 mm, align=center] {};
    \node[text] (c1t) [right=-1.4cm of c1, align=center, text width =10 mm] {\scriptsize{Component 1}};
    \node[text] (io2) [above right=-0.5 cm and -0.35cm of c1, align=center, text width =10 mm] {\scriptsize{I/O}};
    \node[text] (fix1) [below=0.2cm of c1] {\Large{\xmark}};
    
    \node[inner sep=0pt] (crowd1)  [below= 1.0 cm of c1]
    	{\includegraphics[width=.04\textwidth]{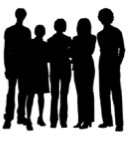}};
    \node[text] (fix1f) [right=0.1cm of crowd1] {\Large{\cmark}};
    
    \node[box] (c2) [right=0.5 cm of c1,
    minimum width = 14 mm, minimum height = 10 mm, text width=7 mm, align=center]{};
    \node[text] (c2t) [right=-1.4cm of c2, align=center, text width =10 mm] {\scriptsize{Component 2}};  
        \node[text] (io3) [above right=-0.5 cm and -0.35cm of c2, align=center, text width =10 mm] {\scriptsize{I/O}};
    
    \node[text] (fix2) [below=0.2cm of c2] {\Large{\xmark}};
    
    \node[inner sep=0pt] (crowd2)  [below= 1.0 cm of c2] {\includegraphics[width=.04\textwidth]{images/crowd.png}};
    \node[text] (fix2f) [right=0.1cm of crowd2] {\Large{\cmark}};
     \node[inner sep=0pt] (caption)  [below right= 0.1 cm and -3.0 cm of crowd2]
    	{Component-based machine learning system};
    \node[inner sep=0pt] (conn2)  [left= 0.2 cm of crowd2]
    	{};
    \node[inner sep=0pt] (conn2box)  [above= 1.3 cm of conn2]
    	{};

    \node[box] (c3) [right=0.5 cm of c2,
    minimum width = 14 mm, minimum height = 10 mm, text width=7 mm, align=center]{};
    \node[text] (c3t) [right=-1.4cm of c3, align=center, text width =10 mm] {\scriptsize{Component 3}};
    
    \node[text] (fix3) [below=0.2cm of c3] {\Large{\xmark}};
    
    \node[inner sep=0pt] (crowd3)  [below= 1.0 cm of c3]
    	{\includegraphics[width=.04\textwidth]{images/crowd.png}};
    \node[text] (fix3f) [right=0.1cm of crowd3] {\Large{\cmark}};
    	
    \node[inner sep=0pt] (conn3)  [left= 0.2 cm of crowd3]
    	{};
    \node[inner sep=0pt] (conn3box)  [above= 1.3 cm of conn3]
    	{};
    \node[inner sep=0pt] (conn4)  [right= 0.6 cm of crowd3]
    	{};
    \node[inner sep=0pt] (conn5)  [above right= 1.5 cm and 0.6 cm of crowd3]
    	{};
        
    \node[inner sep=0pt] (user) [right=1.9 cm of c3, label=above:] 
    {\includegraphics[width=.04\textwidth]{images/crowd.png}};
    \node[inner sep=0pt] (connuser)  [above left= -0.25 cm and 0.0 cm of user]
    	{};
    
    \node[inner sep=0pt] (connuserbox)  [above left= -0.25 cm and 1.3 cm of user]
    	{};

    \node[inner sep=0pt] (database) [below=1.1 cm of user, label=below:Logs]
    	{\includegraphics[width=.04\textwidth]{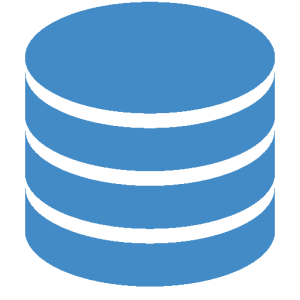}};
    
     \draw [draw=Black!100] (c1) [->] -- (c2);
     \draw [draw=Black!100] (c2) [->] -- (c3);
     \draw [draw=Black!100, densely dashed] (c1) [->] -- (crowd1);
     \draw [draw=Black!100, densely dashed] (c2) [->] -- (crowd2);
     \draw [draw=Black!100, densely dashed] (c3) [->] -- (crowd3);
	 \draw [draw=Black!100, densely dashed] (crowd1) [-] -- (conn2);
     \draw [draw=Black!100, densely dashed] (conn2) [->] -- (conn2box);
     \draw [draw=Black!100, densely dashed] (crowd2) [-] -- (conn3);
     \draw [draw=Black!100, densely dashed] (conn3) [->] -- (conn3box);
     \draw [draw=Black!100, densely dashed] (crowd3) [-] -- (conn4);
     \draw [draw=Black!100, densely dashed] (conn4) [->] -- (conn5);
          
     \draw [draw=Black!100] (c3) [->] -- (user);
     
     \node[text] (response) [below left=-0.4 cm and 0.0cm of user, align=center, text width =10 mm]
     {\scriptsize{}};
     \node[text] (query) [above =0.05 cm of response, align=center, text width =10 mm]
     {\scriptsize{System output}};
	 \node[inner sep=0pt] (arrow) [below=1.4 cm of response] 
    	{\includegraphics[width=.04\textwidth]{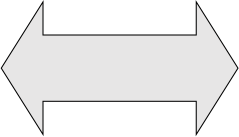}};
     	 \node[text] (log) [below =0.0 cm of arrow, align=center, text width =10 mm]
     {\scriptsize{Failures Fixes}};
     \node[inner sep=0pt] (arrow2) [above left=0.95 cm and -0.55 cm of database, rotate=90] 
    	{\includegraphics[width=.04\textwidth]{images/arrow.png}};
	 \node[text] (eval) [above left =0.9 cm and 0.0 cm of arrow2, align=center, text width =10 mm, rotate=90]
     {\scriptsize{Evaluation}};

\end{tikzpicture}
\caption{Troubleshooting with humans in the loop}
\label{fig:intro}
\end{figure}

{\noindent \bfseries Approach.} We introduce a
troubleshooting methodology which relies on crowdworkers to identify and fix mistakes in existing systems.
Human intervention is crucial to the approach as human fixes simulate improved
component output that cannot be produced otherwise without significant system
development efforts. Figure 1 shows the main flow of our approach. First,
workers evaluate the system output without any fix to analyze the current system
state. To simulate a component fix, the input and output of the component
accompanied with the fix description are sent to a crowdsourcing platform as
microtasks.
Once workers apply the targeted fixes for a component, the fixed output is
integrated back into the running system, which thereby generates an improved
version of the component. The system is executed as a simulation with the injected fix and the output is evaluated again via crowdworkers. The
overall process collects a valuable set of log data on system failures, human
fixes, and their impact on the final output. This data can then be analyzed to identify the most effective combination of component
improvements to guide future development efforts.

{\noindent \bfseries Case study.} We apply our methodology to a state-of-the-art
integrative learning system developed to automatically caption images \cite{fang2015captions}. The system
involves three machine learning components in a pipeline, including \emph{visual
detection}, \emph{language generation}, and \emph{caption ranking}. The multimodal nature of
this case study allows us to demonstrate the applicability of the approach for
components processing different forms of data and carrying out various tasks.
The methodology reveals new insights on the error
dynamics previously unknown to the designers, and offers recommendations on how
to best improve the system. For example, in contrast to what system designers had assumed, improving the
Reranker is more effective than improving the Visual Detector.
Experiments highlight the benefits of making informed decisions about component
fixes as their effectiveness varies greatly (18\%, 3\% and 27\%  for the three
components respectively). 

\section{Background and Problem Characterization}
We now define the problem of troubleshooting component-based
 machine learning systems. We start by
describing the image captioning system as a running example and then
continue with the problem formalization.

\begin{figure}
\centering
\begin{tikzpicture}
	\footnotesize
	\tikzstyle{text}=[draw=none,fill=none]
	\tikzstyle{box}=[rectangle, draw=MidnightBlue!100]
	
    \node[box] (system) [label=below:,
    minimum width = 1.0\columnwidth, minimum height = 26 mm]{};
    
    \node[box] (c1) [above left= -2.5 cm
    and -1.8 cm of system, minimum width = 14 mm, minimum height = 10 mm, text width=15 mm,
    align=center] {Visual Detector};    
    
    \node[text] (w1) [above right=1.05cm and -0.2 cm of c1,
    minimum width = 14 mm, text width=17 mm,
    align=center] {\scriptsize{snowboard, 0.96}};
    \node[text] (w2) [below=-0.15 cm of w1,
    minimum width = 14 mm, text width=17 mm,
    align=center] {\scriptsize{ snow, 0.94}};
    \node[text] (w3) [below=-0.15 cm of w2,
    minimum width = 14 mm, text width=17 mm,
    align=center] {\scriptsize{ man, 0.89}};
    \node[text] (w4) [below=-0.15 cm of w3,
    minimum width = 14 mm, text width=17 mm,
    align=center] {\scriptsize{ mountain, 0.87}};
    \node[text] (w5) [below=-0.15 cm of w4,
    minimum width = 14 mm, text width=17 mm,
    align=center] {\scriptsize{ skis, 0.71}};
    \node[text] (w6) [below=-0.15 cm of w5,
    minimum width = 14 mm, text width=17 mm,
    align=center] {\scriptsize{$\ldots$}};
    \node[text] (io12) [below=-0.07 cm of w6,
    minimum width = 14 mm, text width=37 mm,
    align=center] {I/O};
    
    \node[box] (c2) [right=1.55 cm of c1,
    minimum width = 14 mm, minimum height = 10 mm, text width=15 mm,
    align=center]{Language Model};   
    \node[text] (s1) [above right=1.05cm and -1.2 cm of c2,
    minimum width = 14 mm, text width=37 mm,
    align=center] {\scriptsize{A man flying through}};
    \node[text] (s2) [below=-0.15 cm of s1,
    minimum width = 14 mm, text width=37 mm,
    align=center] {\scriptsize{the air on a snowboard.}};    
    \node[text] (s3) [below=-0.05 cm of s2,
    minimum width = 14 mm, text width=37 mm,
    align=center] {\scriptsize{$\ldots$}};
    \node[text] (s4) [below=-0.05 cm of s3,
    minimum width = 14 mm, text width=37 mm,
    align=center] {\scriptsize{A man riding skis}};
    \node[text] (s5) [below=-0.15 cm of s4,
    minimum width = 14 mm, text width=37 mm,
    align=center] {\scriptsize{on a snowy mountain.}};
    \node[text] (io23) [below=0.0 cm of s5,
    minimum width = 14 mm, text width=37 mm,
    align=center] {I/O};
    
    \node[box] (c3) [right=1.55 cm of c2,
    minimum width = 14 mm, minimum height = 10 mm, text width=15 mm,
    align=center]{Caption Reranker};
    
    \node[inner sep=0pt] (image)  [above= 1.6 cm of c2, draw=gray]
    	{\includegraphics[width=.27\columnwidth]{images/examples/COCO_val2014_000000173245.jpg}};
	
	\draw [draw=Black!100, very thick] (c1) [->] -- (c2);
	\draw [draw=Black!100, very thick] (c2) [->] -- (c3);

    \node[inner sep=0pt] (connleftimage)  [below left= -0.2 and -0.05 of image]
    	{};
    \node[inner sep=0pt] (connleft)  [left= 1.8 of connleftimage]
    	{};
    	
    \node[inner sep=0pt] (connleft2)  [left= 1.75 of connleftimage]
    	{};
    \node[inner sep=0pt] (connleftc1)  [below left=1.6cm and 1.75 of image]
    	{};
	\draw [draw=Black!100, very thick] (connleftimage) [-] -- (connleft);
	\draw [draw=Black!100, very thick] (connleft2) [->] -- (connleftc1);

	\node[inner sep=0pt] (connrightimage)  [below right= -0.2 and -0.05 of image]
    	{};
    \node[inner sep=0pt] (connright)  [right= 1.8 of connrightimage]
    	{};
    	
    \node[inner sep=0pt] (connright2)  [right= 1.75 of connrightimage]
    	{};
    \node[inner sep=0pt] (connrightc1)  [below right=1.6cm and 1.75 of image]
    	{};
	\draw [draw=Black!100, very thick] (connrightimage) [-] -- (connright);
	\draw [draw=Black!100, very thick] (connright2) [->] -- (connrightc1);
	\node[rectangle, fill=white] (caption) [above=2.3 cm of c3,
    minimum width = 14 mm, minimum height = 10 mm, text width=27 mm,
    align=center]{\#1 \\A man flying through the air on a snowboard.};  
	\draw [draw=Black!100, very thick] (c3) [->] -- (caption);
		
\end{tikzpicture}
\caption{The image captioning system}
\label{fig:captionsystem}
\end{figure}
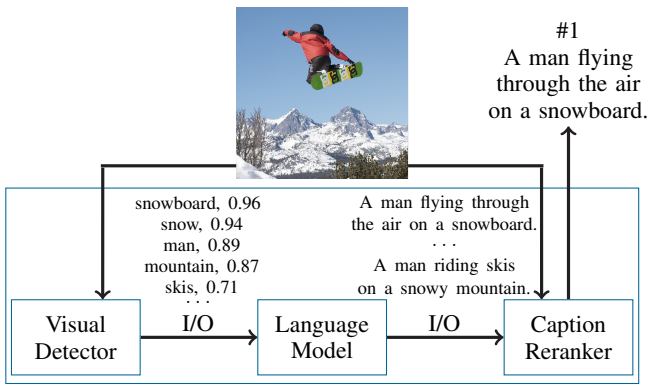

\subsection{Case Study: An Image Captioning System}
The system \cite{fang2015captions} that we use as a case study automatically
generates captions as textual descriptions for images. This task has emerged as
a challenge problem\footnote{http://mscoco.org/dataset/\#captions-challenge2015}
in artificial intelligence as it involves visual and language
understanding and has multiple real-world applications, including the provision
of descriptions for assisting visually impaired people.
Figure~\ref{fig:captionsystem} shows the system architecture, consisting of
three machine learning components. The first and third components leverage
convolutional neural networks combined with multiple instance learning
\cite{zhang2005multiple}. The second one is a maximum-entropy language
model \cite{berger1996maximum}.

\noindent{\bfseries{Visual Detector.}} The first component takes an image as an
input and detects a list of words associated with recognition scores. The
detector recognizes only a restricted vocabulary of the 1000 most common words 
in the training captions.

\noindent{\bfseries{Language Model.}} 
This component is a statistical model that generates likely word sequences as
captions based on the words recognized from the Visual Detector, without having
access to the input image. The set of the 500 most likely image captions and the
respective log-likelihood scores are forwarded to the Caption Reranker.

\noindent{\bfseries{Caption Reranker.}} 
The task of the component is to rerank the captions generated from the
Language Model and select the best match for the image. The
Reranker uses multiple features among which the similarity between the
vector representations of images and captions. The caption with the highest
ranking score is selected as the final best caption.

\noindent{\bfseries{Dataset.}} All components are individually trained on the
MSCOCO dataset \cite{lin2014microsoft} which was built as an image
captioning training set and benchmark.
It contains 160,000 images as well as five human-generated
captions for all images. We use images randomly sampled from the validation dataset to evaluate
our approach.
 
\subsection{Problem Characterization}
\noindent{\bfseries{Problem context.}} This work studies machine learning
systems consisting of several components designed to carry out specific tasks in
the system. The system takes a set of data as \emph{system input} and the
individual components work together to produce a final \emph{system output}. We
assume that the system architecture is provided to the methodology by system designers by specifying:

\begin{my_enumerate}
\item The
set of system components along with the \emph{component
input} and \emph{output} data types.
\item A set of directed communication dependencies between components
  denoting the input / output exchange. The whole set of dependencies defines
  the system \emph{execution workflow}. In our methodology, we only handle acyclic dependencies but allow for branching. 
\end{my_enumerate}

\noindent{\bfseries{Problem definition.}} Troubleshooting of component-based machine learning systems can be decoupled to answering the following two
questions:

\noindent\fbox{\begin{minipage}{0.97\columnwidth} {\bf Question 1:} \emph{How does the system fail?}
\\
The system designer is interested in identifying and measuring the different types of \emph{system failures} and their frequencies as well as the failures of individual components in the system context. 
\end{minipage}}

\noindent \fbox{\begin{minipage}{0.97\columnwidth}
{\bf Question 2:} \emph{How to improve the system?}
\\
System failures can be addressed by various potential fixes applicable to
individual components. To guide future efforts on improving the system, the
system designer is interested in knowing the effects of component fixes on: (i)
specific input instances, and (ii) the overall system output quality. The first
requirement is relevant to fine-grained instance-based debugging that a
troubleshooting methodology should implement, while the second one provides guidance on future efforts to improve the whole system.

\end{minipage}}
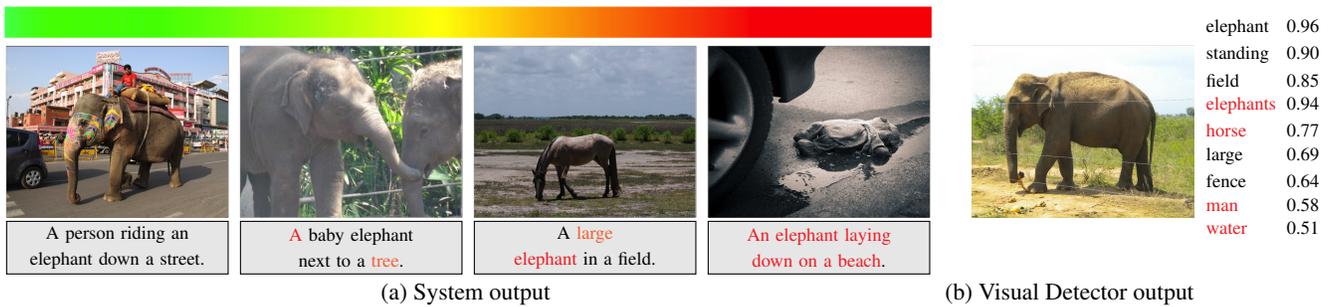
\begin{figure*}[t!]
\centering
\begin{tikzpicture}
	\tikzstyle{text}=[draw=none,fill=none]
	\definecolor{light-gray}{gray}{0.9}
	\tikzstyle{box}=[rectangle, fill=light-gray,minimum width = .35\columnwidth,
	text width = .35\columnwidth, align=center, minimum height = 7
	mm,font=\scriptsize, inner sep=0,outer sep=0, draw=black]
    \node[inner sep=0pt] (spectre)  []
    	{\includegraphics[width=1.465\columnwidth]{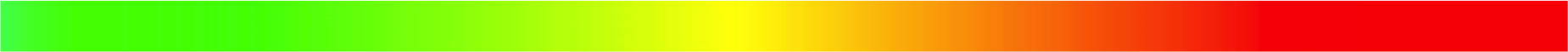}};
	\node[inner sep=0pt] (el1)  [below left= 0.1 cm and -2.99 cm of spectre,
	draw=gray] {\includegraphics[width=.35\columnwidth, height=.27\columnwidth]{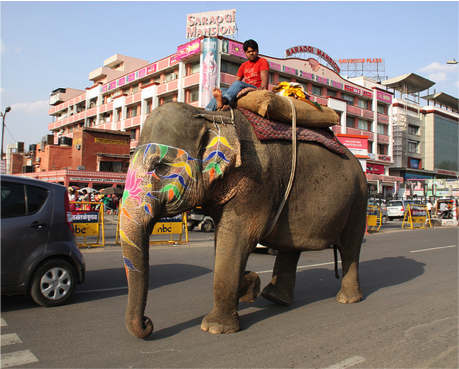}};
	\node[inner sep=0pt] (el2)  [right= 0.15 cm of el1, draw=gray]
    	{\includegraphics[width=.35\columnwidth, height=.27\columnwidth]{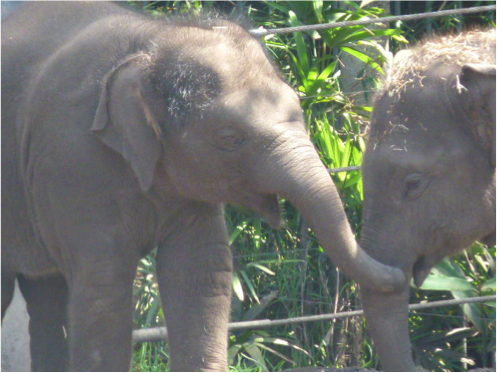}};
    \node[inner sep=0pt] (el3)  [right= 0.15 cm of el2, draw=gray]
    	{\includegraphics[width=.35\columnwidth, height=.27\columnwidth]{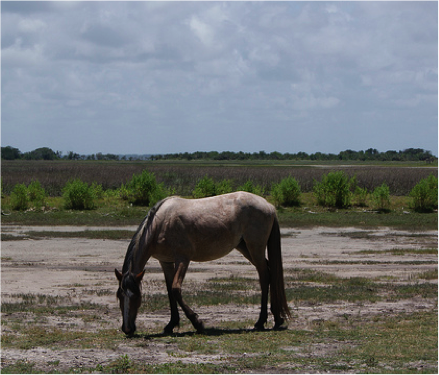}};
    \node[inner sep=0pt] (el4)  [right= 0.15 cm of el3, draw=gray]
    	{\includegraphics[width=.35\columnwidth, height=.27\columnwidth]{images/examples/el4.png}};
     
    \node[box] (c1) [below=0.05cm of el1]{A person riding an\\elephant down a
    street.}; \node[box] (c2) [below=0.05cm of el2]{\textcolor{Red}{A} baby
    elephant next to a \textcolor{RedOrange}{tree}.}; \node[box] (c3) [below=0.05cm of el3]{A \textcolor{RedOrange}{large} \\ \textcolor{Red}{elephant} in a field.};
        \node[box] (c4) [below=0.05cm of el4]{\textcolor{Red}{An elephant laying down on a beach}.};
    
    \node[inner sep=0pt] (el5)  [right= 0.55 cm of el4, draw=gray]
    	{\includegraphics[width=.35\columnwidth,
    	height=.27\columnwidth]{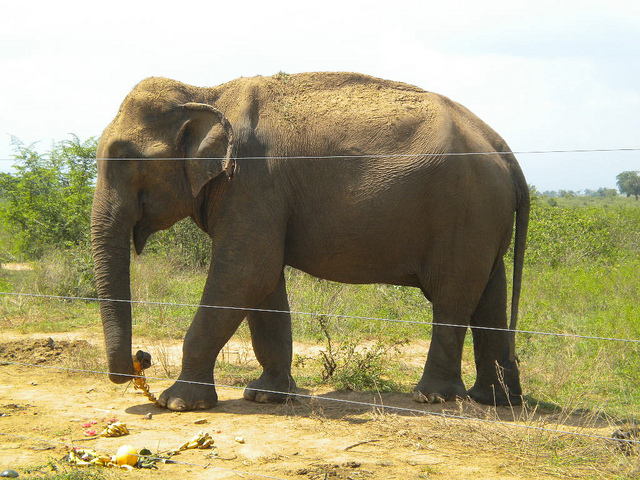}};
    	 \matrix[ 
   above  right=0.5 cm and 1.0 cm of el5,
	matrix of nodes,
    anchor=north, 
    draw = none,
    inner sep=0.2em,
    font=\scriptsize ,
    column 1/.style={anchor=base west},
    column 2/.style={anchor=base west},
    column 3/.style={anchor=base west},
    column 4/.style={anchor=base west},
  ] (wordsdetector)
  {
    elephant	& 0.96	&[5pt]\\
    standing	& 0.90	&[5pt]\\
    field		& 0.85 	&[5pt]\\
    \textcolor{Red}{elephants}	& 0.94 	&[5pt]\\
    \textcolor{Red}{horse}		& 0.77 	&[5pt]\\
    large 						& 0.69 	&[5pt]\\
	fence 						& 0.64 	&[5pt]\\
	\textcolor{Red}{man} 		& 0.58 	&[5pt]\\
	\textcolor{Red}{water} 		& 0.51 	&[5pt]\\
    };

	\node[text] (a) [below right=0cm and -1.2 cm of c2, align=center,
	font=\small]{(a) System output};
    \node[text] (b) [right=5.0cm of a, align=center, font=\small]{(b) Visual
    Detector output};
    
\end{tikzpicture}
\caption{Continuous output quality in the captioning system}
\label{fig:continuous}
\end{figure*}
\begin{figure}[t!]
\centering
\begin{tikzpicture}
    \tikzstyle{text}=[draw=none,fill=none]
	\definecolor{light-gray}{gray}{0.9}
	\tikzstyle{boxcaption}=[rectangle, inner sep=0.5 mm, fill=light-gray,minimum width =
	.25\columnwidth, text width = .16\columnwidth, align=center, minimum height = 6 mm, font=\scriptsize]
    \node[inner sep=0pt] (blender)  [draw=gray]
    	{\includegraphics[width=.35\columnwidth,height=.49\columnwidth]{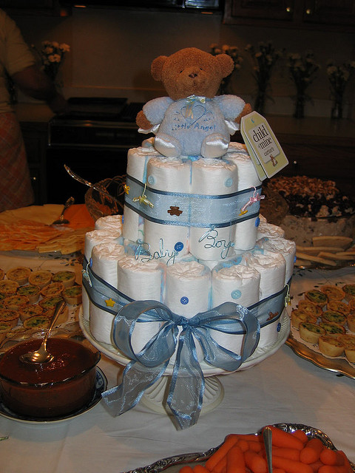}};
   \node[text] (c1t) [above right=-0.7 cm and 0.0 cm of blender, align=center,
    text width = 17 mm]{\bfseries{Visual Detector}};
   \node[text] (c2t) [above right=-0.86 cm and 0.0 cm of c1t, align=center,
    text width = 17 mm]{\bfseries{Language Model}};  
   \node[text] (c3t) [above right=-0.86 cm and 0.3 cm of c2t, align=center,
    text width = 17 mm]{\bfseries{Caption Reranker}};
    
    \matrix[ 
    below=-0.1 cm of c1t,
	matrix of nodes,
    anchor=north, 
    draw = none,
    inner sep=0.2em,
    font=\scriptsize ,
    column 1/.style={anchor=base west},
    column 2/.style={anchor=base west},
    column 3/.style={anchor=base west},
    column 4/.style={anchor=base west},
  ] 
  {
    1. teddy 	& 0.92 	&[5pt]\\
    2. on	& 0.92 	&[5pt]\\
    3. cake	& 0.90 	&[5pt]\\
    4. bear	& 0.87 	&[5pt]\\
    5. stuffed	& 0.85 	&[5pt]\\    
    \quad \dots	&  	&[5pt]\\
    15. \textcolor{Red}{blender}	& 0.57 	&[5pt]\\};
    \node[boxcaption, draw=black] (caption1) [below=-0.02 cm of c3t]{1. A \textcolor{Red}{blender} sitting on top of a
    cake.}; \node[boxcaption] (caption2) [below=0.05 cm of caption1]{2. A teddy bear \textcolor{Red}{in front of} a birthday cake.};
    \node[boxcaption] (caption3) [below=0.05 cm of caption2]{3. A cake sitting on top of a \textcolor{Red}{blender}.};
    
    \node[boxcaption, draw=black] (caption1l) [below=-0.02 cm of c2t]{1. A teddy \\ bear.};
    \node[boxcaption] (caption2l) [below=0.05 cm of caption1l]{2. A stuffed \\ bear.};
    \node[text] (dot) [below=0.2 cm of caption2l, align=center,
    text width = 17 mm]{\ldots};
    
    \node[boxcaption] (caption3l) [below=0.05 cm of dot]{108. A \textcolor{Red}{blender} sitting on top of a
    cake.};

\end{tikzpicture}
\caption{Component entanglement in the captioning system}
\label{fig:entanglement}
\end{figure}
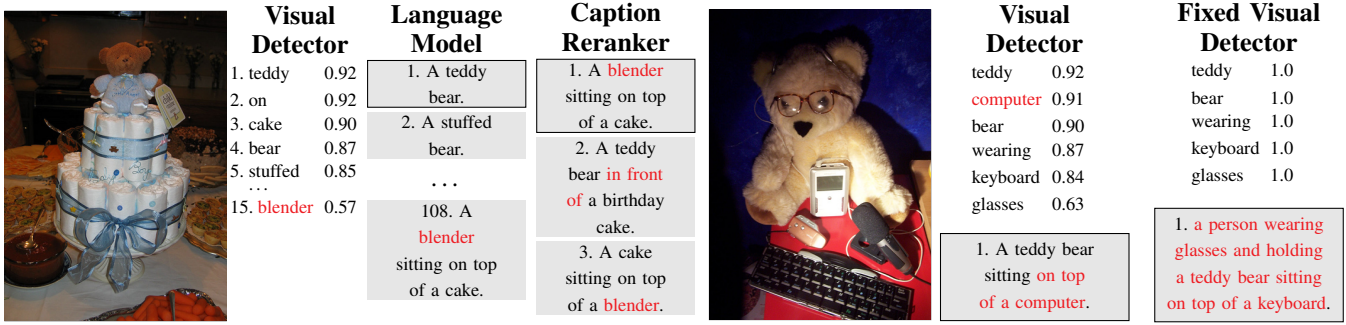

\noindent The first question aims at analyzing the current state of the system.
The second question explores future opportunities and investigates the
efficiency of different strategies to improve the system. Next, we examine the special
characteristics of component-based machine learning systems that make the problem of troubleshooting challenging. These
characteristics differentiate this problem from previous work on troubleshooting
and motivate our methodology.

\noindent{\bfseries{Continuous quality measures.}} Uncertainty is inherent in
 machine learning components. When these components work
together to solve complex tasks, the measure of quality for individual
components and the system as a whole is no longer binary, rather it spans a wide spectrum. Therefore, the evaluation
of these systems needs to go beyond accuracy metrics to deeper analysis of system
behavior. For instance, Figure~\ref{fig:continuous} shows a few examples from
image captioning where the system and component output varies in quality and the
types of mistakes. Troubleshooting in this quality continuum where all
components are only partially correct is non-trivial.

\noindent{\bfseries{Complex component entanglement.}} In component-based machine learning systems, components
have complex influences on each other as they may be tightly coupled or the boundaries between their responsibilities
may not be clear. When the quality of a component depends on the output of previous components, blame cannot be
assigned to individual components without decoupling imperfection problems in
component inputs. Figure~\ref{fig:entanglement} illustrates a typical scenario of component entanglement in the image captioning system.
The final caption is clearly unsatisfactory as it mentions a non-existing
object (\texttt{blender}).
However, the Visual Detector
assigns a low score to the word \texttt{blender} (0.57), which makes the
detector only partially responsible for the mistake. The Language Model is also partially responsible as it creates a
sentence with low commonsense awareness. Finally, the caption reranker chooses
as the best caption a sentence that includes a word with a low score. In this
example, errors from all components are interleaved and it is not possible
to disentangle their individual impact on the final error.

\begin{figure}[t!]
\centering
\begin{tikzpicture}
    \tikzstyle{text}=[draw=none,fill=none]
	\definecolor{light-gray}{gray}{0.9}
	\tikzstyle{boxcaption}=[rectangle, fill=light-gray,minimum width =
	.25\columnwidth, text width = .27\columnwidth, align=center, minimum height = 6 mm, font=\scriptsize]
    \node[inner sep=0pt] (blender)  [draw=gray]
    	{\includegraphics[width=.35\columnwidth,height=.49\columnwidth]{images/examples/COCO_val2014_000000042889.jpg}};
   \node[text] (c1t) [above right=-0.7 cm and 0.4 cm of blender, align=center,
    text width = 17 mm]{\bfseries{Visual Detector}};
   \node[text] (c2t) [above right=-0.86 cm and 0.4 cm of c1t, align=center,
    text width = 27 mm]{\bfseries{Fixed Visual Detector}};
    
    \matrix[ 
    below=-0.1 cm of c1t,
	matrix of nodes,
    anchor=north, 
    draw = none,
    inner sep=0.2em,
    font=\scriptsize ,
    column 1/.style={anchor=base west},
    column 2/.style={anchor=base west},
    column 3/.style={anchor=base west},
    column 4/.style={anchor=base west},
  ] (wordsdetector)
  {
    teddy	& 0.92 	&[5pt]\\
    \textcolor{Red}{computer}	& 0.91 	&[5pt]\\
    bear	& 0.90 	&[5pt]\\
    wearing	& 0.87 	&[5pt]\\
    keyboard	& 0.84 	&[5pt]\\
    glasses	& 0.63 	&[5pt]\\};
    \node[boxcaption, draw=black] (caption1) [below=0.1 cm of wordsdetector]{1. A teddy bear sitting \textcolor{Red}{on top of a computer}.};
    \matrix[ 
    below=-0.1 cm of c2t,
	matrix of nodes,
    anchor=north, 
    draw = none,
    inner sep=0.2em,
    font=\scriptsize ,
    column 1/.style={anchor=base west},
    column 2/.style={anchor=base west},
    column 3/.style={anchor=base west},
    column 4/.style={anchor=base west},
  ] (wordsdetectorfixed)
  { 
    teddy	& 1.0 	&[5pt]\\
    bear	& 1.0 	&[5pt]\\
    wearing	& 1.0 	&[5pt]\\
    keyboard	& 1.0 	&[5pt]\\
    glasses	& 1.0 	&[5pt]\\};
     \node[boxcaption, draw=black] (caption2) [below=0.15 cm of wordsdetectorfixed]{1. \textcolor{Red}{a person wearing glasses and holding 
a teddy bear sitting on top of a keyboard}.};
\end{tikzpicture}
\caption{Non-monotonic error in the captioning system}
\label{fig:nonmonerror}
\end{figure}
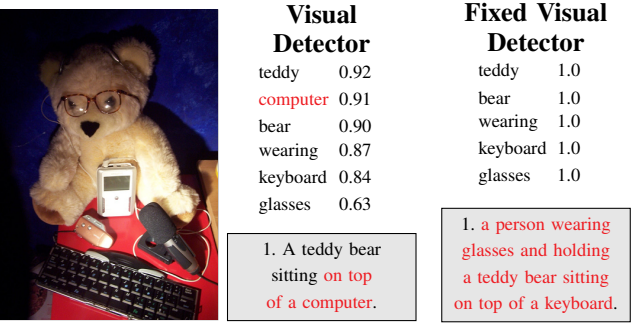

\noindent{\bfseries{Non-monotonic error.}} 
We note that improving the outputs of components does not guarantee system
improvement.
On the contrary, doing so may lead to quality deterioration.
For example, when components are tuned to suppress erroneous
behavior of preceding components, applying fixes to the earlier ones may result
to unknown failures.
Figure~\ref{fig:nonmonerror} shows an example of non-monotonic error behavior.
Here, the Visual Detector makes a mistake by including \texttt{computer} in the
list. The initial assumption would be that if the list is fixed so that it
contains only the prominent words, then the quality of the caption should
increase. In reality, the caption after the fix is more erroneous than the
original. Since the language model finds a teddy bear wearing glasses unlikely,
it creates a caption that mentions a \texttt{person} instead.

\section{Human-in-the-loop Methodology}
Due to these problem characteristics, blame assignment is challenging in integrative systems and analyzing only the
current state of the system is not sufficient to develop strategies for system improvement. As shown in
Figure~\ref{fig:intro}, our methodology overcomes these challenges by introducing humans in the loop for: (i) simulating
component fixes, and (ii) evaluating the system before and after fixes to
directly measure the effect of future system improvements.
%
%

\noindent{\bfseries{Methodology setup.}} 
The troubleshooting methodology is applicable to systems that follow the assumptions of:
(i) \emph{system modularity} with clearly defined component inputs and outputs,
and (ii) \emph{human interpretability} of the component input / output. To
apply the methodology to a new system, the system designer provides the set of components and their input/outputs within the system execution workflow. After
identifying a list of component fixes that can potentially improve the system, the system designer formulates
corresponding crowdsourcing tasks for these fixes and the overall system evaluation. Both types of tasks should describe
the high-level goal of the system, the context in which it operates as well as its requirements (\emph{e.g.} an image
captioning system designed to assist users with visual impairments).
In addition, component fixing tasks should be appropriately formulated so that their expected output matches the output
of implementable fixes that the system designer plans to test.

\noindent{\bfseries{Troubleshooting steps.}} The execution of the methodology is guided by the \emph{fix workflow},
which is a combination of various component fixes to be evaluated. The system designer chooses which fix
workflows to execute and evaluate for the purpose of troubleshooting. For a given fix workflow, the steps of our methodology are as follows:

\begin{my_enumerate}
  \item \emph{Current system evaluation} --- workers assess the final output of the current system on various quality
  measures. 
  \item \emph{Component fix simulation} --- for each fix in the workflow,
  workers complete the respective micro-task for examining and correcting the
  component output.
  \item \emph{Fix workflow execution} --- executing a fix workflow involves integrating the corrected
  outputs of each component into the system execution.
  \item \emph{After-fix system evaluation} --- workers re-evaluate the new 
  system output after incorporating component fixes.
\end{my_enumerate}

When a workflow includes fixes for multiple components, steps 2 and 3 need to be repeated so that the fixes of earlier
components are reflected on the inputs of later components. 

\noindent{\bfseries{Troubleshooting outcomes.}} 
Applying human fix workflows simulates improved component states and helps
system designers to observe the effect of component fixes on system performance, overcoming the challenges raised by the
problem characteristics. 
\begin{my_enumerate}
\item \emph{Continuous quality measures} --- Comparing
the system quality before and after various fix workflow executions 
not only can quantify the current quality of system and component output, but it
can also isolate and quantify the effect of individual component fixes. For example,
if many components are partially failing and are possibly responsible for a
specific error, the system designer can test the respective fixes, systematically understand their impact, and
decide which are the most promising ones.
\item \emph{Non-monotonic error} --- Non-monotonic error propagation
can be disclosed when the overall system quality drops after a component fix.
When such a behavior is observed, the system designer can conclude that although
these fixes may improve the internal component state, they are not
advisable to be implemented in the current state of the system as they produce
negative artifacts in the holistic system. 
\item \emph{Complex component entanglement} --- Entanglement detection requires
the execution of workflows with different combinations of fixes to measure the individual and the joint effect
of component fixes. For example, if two consecutive components 
are entangled, individual fixes in either one of them may not improve
the final output. However, if both components are fixed jointly,
this may trigger a significant improvement. The designer could also use this information
to detect entanglement and potentially correct the system architecture in future
versions.
\end{my_enumerate}

\section{Troubleshooting the Image Captioning System}
We now describe the customized crowdsourcing tasks for our case
study for both system evaluation and component fixes. 
Table~\ref{tb:fix} lists
all component fixes specifically designed for this case study. The task design
is an iterative process in collaboration with system designers so that 
human fixes can appropriately simulate implementable improvements.

\begin{table} 
\centering
\footnotesize
\begin{tabular}{l l} 
  \Xhline{1pt}
  Component & Fix description \\ \Xhline{1pt}
  Visual Detector & Add / remove objects \\
  Visual Detector & Add / remove activities  \\ \hline
  Language Model & Remove noncommonsense captions  \\
  Language Model & Remove non-fluent captions  \\ \hline
  Caption Reranker & Rerank Top 10 captions  \\ 
   \Xhline{1pt}
    \end{tabular}
    \caption{Summary of fixes for the image captioning system.}
    \label{tb:fix}
\end{table}

\noindent{\bfseries System evaluation.}
The system evaluation task is designed to measure different quality metrics associated with captions as well as the overall
\emph{human satisfaction}. The task shows workers an
image-caption pair and asks them to evaluate the following
quality measures: \emph{accuracy} (1-5 Likert scale), \emph{detail} (1-5 Likert scale),
\emph{language} (1-5 Likert scale), \emph{commonsense} (0-1), and \emph{general
evaluation}. For each measure, we provided a detailed description along with
representative examples. However, we intentionally did not instruct workers for
the general evaluation to prevent biasing 
them on which quality measure is more important (\emph{e.g.} accuracy vs. detail).
 
{\noindent\bfseries{Visual Detector fixes.}} 
We designed two different tasks for
fixing object and activity detections respectively represented by nouns and verbs in the word list.
The \emph{object fix} shows workers the input image with the
list of nouns present in the visual detector output. Workers are asked to
correct the list by either removing objects or adding new ones. 
The \emph{activity fix} has a similar design. 
We intentionally group \emph{addition} and \emph{removal} fixes to simulate
improvements of the Visual Detector in precision and recall as potential
implementable fixes in this component.
The result of any Visual Detector fix is a new word list which is passed to the
Language Model together with the worker agreement scores (\emph{e.g.} majority vote). 

{\noindent\bfseries{Language Model fixes.}} These fixes are designed for
removing sentences that are either not commonsense or not fluent. The
tasks do not share the input image with the worker, as the
Language Model itself does not have access to the image.
In the \emph{commonsense fix}, workers mark whether a caption describes a likely
situation that makes sense in the real world. For example, the caption \texttt{A cat playing a video game} has no commonsense awareness in a general context. In
the \emph{language fix}, the goal is to evaluate the language fluency of captions in a 1-5
Likert scale. In addition, workers highlight problematic parts of the sentence which they
think would make the caption fluent if fixed appropriately. The resulting list
of problematic segments is a reusable resource to filter out captions that
contain the same patterns.

Both fixes simulate improved versions of the Language Model that generate
commonsense sentences in a fluent language. To
integrate Language Model fixes in the system execution we exclude the
noncommonsense and the non-fluent captions from the list forwarded to the
Caption Reranker.

{\noindent\bfseries{Caption Reranker fixes.}} This task shows an image together with the corresponding top 10 captions from the Reranker
(in random order), and asks workers to pick up to 3 captions that they think fit
the image best. The answers are then aggregated via majority vote and the caption with the highest agreement is selected as the new system output.

\section{Experimental Evaluation}
The evaluation of the captioning system with our methodology uses an
\eval\ dataset of 1000 images randomly selected from the MSCOCO validation
dataset. All experiments were performed on
Amazon Mechanical Turk. We report the system quality based on human assessments.
An additional analysis using automatic machine translation scores can be found
in the appendix.
\begin{table}
\centering
\footnotesize
\begin{filecontents*}{data/human_current_state.csv}
metric,evaluation_dataset,satisfactory,unsatisfactory
Accuracy (1-5),3.674,4.474,2.579
Detail (1-5),3.563,4.265,2.601
Language (1-5),4.509,4.693,4.256
Commonsense (0-1),0.957,1.000,0.898
General (1-5),3.517,4.306,2.437
\%Satisfactory (0-1),57.8\%,100\%,0\%
\end{filecontents*}
\pgfplotstabletypeset[
    col sep=comma,
    string type,
    every head row/.style={%
	    before row=\Xhline{1pt},
        after row=\Xhline{1pt}
    },
    every last row/.style={after row=\hline},    
    columns/metric/.style={column name=, column type={p{2.7cm}|}},
    columns/evaluation_dataset/.style={column name=\evalshort, column type={c|}},    
    columns/satisfactory/.style={column name=\satshort, column type={c|}},
    columns/unsatisfactory/.style={column name=\unsatshort, column type={c}},    
    every last row/.style={after row=\Xhline{1pt}} 
    ]{data/human_current_state.csv} 
\caption{Current system evaluation.}
\label{tb:current}
\end{table}

\subsection{Current System State}
First, we evaluate the current system state as shown in Table~\ref{tb:current}. To gain a
deeper understanding of the system performance, we divide the \eval\ data in two datasets: \sat\ and \unsat\ based on the
general evaluation score collected from the system evaluation task. We consider every answer in 1-3 as an
unsatisfactory evaluation, and every other answer in 4-5 as satisfactory. All
instances whose original caption reaches a majority agreement on
being satisfactory belong to the \sat\ dataset. The
rest is classified as \unsat. 

\noindent\fbox{\begin{minipage}{0.97\columnwidth} {\bf Result:} Only 57.8\% of the
images in the \eval\ dataset have a satisfactory caption. The
comparison between the \sat\ and \unsat\ partitions shows that the highest discrepancies
happen for the \emph{accuracy} and \emph{detail} measures, highlighting the correlation of accuracy and detail
with the overall satisfaction.
\end{minipage}}

\subsection{Current Component State}
An additional functionality of a human-assisted troubleshooting methodology is to 
evaluate the quality of the existing individual system
components. 
\begin{figure}[t]
\footnotesize
\begin{tikzpicture}
        \begin{axis}[
          ybar=.3cm,
          ymin=0,
          ymax=1.1,
          tickpos=left,
          width=0.55\columnwidth,
    	  ylabel shift = -0.1cm,
    	  height=.53\columnwidth,
          xticklabels={Precision, Recall},
          ylabel=Precision / Recall,
          enlarge x limits=0.5,
          legend style={at={(0.05,0.85)},
          anchor=west,legend columns=-1},
          nodes near coords,
          nodes near coords align={vertical},
          xtick=data,
          name=plot1
         ]            
\addplot[fill=blue!25] coordinates {(1, 0.44) (2, 0.92)};
\addplot[fill=green!25] coordinates {(1, 0.8) (2, 0.87)};    
\end{axis}
\hspace{3mm}
\begin{axis}[
  ybar=.3cm,
  legend columns=2, 
  legend style={nodes=right,font=\tiny,anchor=south,at={(0.56,1.0)},
  draw=none, fill=none}, 
  ymin=0,
  ymax=15,
  tickpos=left,
  width=0.55\columnwidth,
  ylabel shift = -0.1cm,
  height=.53\columnwidth,
  xticklabels={Before fix, After fix},
  ylabel=List length,
  enlarge x limits=0.5,
  legend style={at={(0.05,0.85)},
  anchor=west,legend columns=-1},
  nodes near coords,
  nodes near coords align={vertical},
  xtick=data,
  at=(plot1.right of south east), anchor=left of south west
 ]            
\addplot[fill=blue!25] coordinates {(1, 12.2) (2, 5.7)};\label{plot:obj}
\addplot[fill=green!25] coordinates {(1, 2.6) (2, 2.3)};\label{plot:act}
\end{axis}
\matrix[
	matrix of nodes,
    anchor=south,
    draw = none,
    inner sep=0.2em,
    font=\footnotesize,
    column 1/.style={anchor=base west},
    column 2/.style={anchor=base west}
  ]at(6.2cm, 1.8cm) 
  {
    \ref{plot:obj} 	& Objects 		&[5pt]\\
    \ref{plot:act}	& Activities 	&[5pt]\\};
\end{tikzpicture}
\caption{Visual Detector fixes and the component state.}
\label{fig:c1state}
\end{figure}
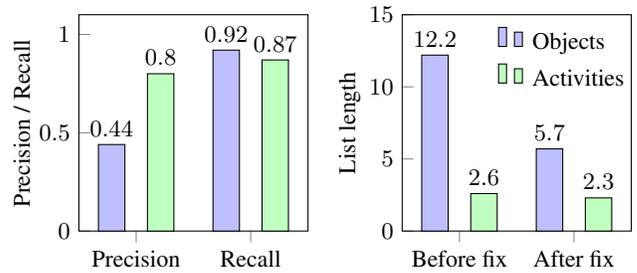
\begin{table}[t]
\centering
\footnotesize
\newcolumntype{C}{>{\centering\arraybackslash}m{1.9cm}}
\newcolumntype{D}{>{\centering\arraybackslash}m{1.4cm}}
\begin{filecontents*}{data/c2_state.csv}
ds,Commonsense Fix,Language Fix,Both Fixes
Top1-\evalshort,8.0\%,22.9\%,\bf{25.0\%}
Top10-\evalshort,8.6\%,21.7\%,\bf{27.1\%}
Top1-\valshort\ ,3.0\%,15.2\%,\bf{16.1\%}
Top10-\valshort\ ,2.6\%,14.2\%,\bf{14.9\%}
\end{filecontents*}
\pgfplotstabletypeset[ 
    col sep=comma,
    string type,
    every head row/.style={before row={\Xhline{1pt}},after row=\Xhline{1pt}},
    every last row/.style={after row=\hline},    
    columns/ds/.style={column name=, column type={p{1.9cm}|}}, 
    columns/Commonsense Fix/.style={column name=Commonsense fix, column type={C|}},
    columns/Language Fix/.style={column name=Language fix, column type={D|}},    
    columns/Both Fixes/.style={column name=Both fixes, column type={D}},
    every last row/.style={after row=\Xhline{1pt}}
    ]{data/c2_state.csv} 
\caption{Percentage of pruned captions from the Language Model and the
component state.}
\label{tb:c2state}
\end{table} 

\noindent {\bfseries Visual Detector} Figure~\ref{fig:c1state} shows the
precision and recall of the Visual Detector for both objects and activities when
compared to the human-curated lists created from the majority vote aggregation of multiple
workers' fixes. In the same figure, we also show the
size of the lists before and after applying human fixes. 

\noindent\fbox{\begin{minipage}{0.97\columnwidth} {\bf Result:} The Visual
Detector produces longer lists for objects than for activities but with
 lower precision and recall. 
\end{minipage}}

\noindent {\bfseries Language Model.} In the Language Model fixes, we examine
only those captions from the Language Model that are among the Top10 best captions in the Caption Reranker. 
 Given that many of the 500 generated sentences never appear as best captions of the Reranker, and the
Language Model output is quite extensive to be fully fixed via crowdsourcing, we focus only on those captions that are
likely to impact the system output. Table~\ref{tb:c2state} shows the percentage of captions pruned after applying
the two fixes. The \val\ dataset here represents the whole MSCOCO dataset which
contains 40,504 images.

\noindent\fbox{\begin{minipage}{0.97\columnwidth} {\bf Result:} Due to self-repetition within the dataset, fixes
generated for the 1000 images of the \eval\ dataset generalize well to the whole
\val\ set, pruning 16.1\% of the Top1 captions and 14.9\%
of the Top10 captions. Language fixes have a higher coverage than the commonsense ones.
\end{minipage}}

\noindent {\bfseries Caption Reranker.} Caption Reranker fixes also focus only on the Top10 best captions. After reranking this set with the crowdsourcing majority vote, we observe that the best caption changes for
76.9\% of the images.
In 46.1\% of the cases, the original caption was never chosen by any of the workers.
For 19.2\% of the images, the majority of workers reported that they could not find any caption in the list that
is a good fit for the image. These cases are indeed more serious failures that
cannot be recovered through reranking only.

\subsection{Component Fixes and System Improvement}
\begin{table}
\centering
\footnotesize
\newcolumntype{C}{>{\centering\arraybackslash}m{0.9cm}}
\newcolumntype{D}{>{\centering\arraybackslash}m{1.0cm}}
\begin{filecontents*}{data/c1_fix_human.csv}
metric,validation,C1_12,C1_34,C1_13,C1_24,C1_1234
Accuracy,3.674,\bfseries{4.045},3.681,3.709,4.000,4.035
Detail,3.563,3.900,3.590,3.604,3.880,\bfseries{3.916}
Language,\bfseries{4.509},4.505,4.427,4.521,4.423,4.432
Csense.,0.957,0.947,0.940,\bfseries{0.957},0.933,0.942
General,3.517,\bfseries{3.848},3.510,3.549,3.796,3.831
\%Sat.,57.8\%,\bfseries{69.1\%},57.1\%,58.5\%,66.8\%,68.0\%
\end{filecontents*}
\pgfplotstabletypeset[
    col sep=comma,
    string type, 
    every head row/.style={%
        before row=\Xhline{1pt},
        after row=\Xhline{1pt}
    }, 
    columns/metric/.style={column name={}, column type={@{\hskip3pt}p{1.1cm}|}},
    columns/validation/.style={column name=No fix, column
    type={@{\hskip3pt}C@{\hskip3pt}|}}, 
    columns/C1_12/.style={column name=Object, column
    type={@{\hskip3pt}C@{\hskip3pt}|}}, columns/C1_34/.style={column name=Activity, column
    type={@{\hskip3pt}C|}}, columns/C1_13/.style={column name=Addition,
    column type={@{\hskip3pt}D|}},  
    columns/C1_24/.style={column name=Removal, column
    type={@{\hskip3pt}D|}}, columns/C1_1234/.style={column name=All fixes,
    column type={@{\hskip3pt}C@{\hskip3pt}}}, every last row/.style={after
    row=\Xhline{1pt}} ]{data/c1_fix_human.csv}
\caption{Visual Detector fixes --- \eval\ dataset.}
\label{tb:fixc1} 
\end{table} 
\noindent{\bfseries{Visual Detector fixes.}} Table~\ref{tb:fixc1} shows
results from applying the four types of fixes on the Visual Detector. These fixes increase the number of satisfactory
captions in the dataset up to 17.6\% compared to the
initial state of the system. Object fixes are more effective than the activity
ones for two reasons. First, the precision of the Visual Detector is originally
significantly lower for objects than for activities (0.44 vs. 0.8), which offers more room for improvement for object
fixes. Second,
activity fixes are limited by the shortcomings of the Language Model. Even when a corrected list of activities is
provided to the Language Model, it may fail to form commonsense captions containing the corrected activities
(\emph{e.g.} \texttt{A woman holding an office}) due to non-monotonic error behavior of the system. 

\noindent\fbox{\begin{minipage}{0.97\columnwidth} {\bf Result:} The entangled
design between the Visual Detector and the Language Model causes non-monotonic
error propagation in particular for activity fixes.
\end{minipage}}
\begin{table}[t]
\centering
\footnotesize
\newcolumntype{C}{>{\centering\arraybackslash}m{1.9cm}}
\newcolumntype{S}{>{\centering\arraybackslash}m{0.9cm}}
\newcolumntype{M}{>{\centering\arraybackslash}m{1.5cm}}

\begin{filecontents*}{data/c2_fix_human.csv}
metric,validation,C2_1,C2_2,C2_12
Accuracy,3.674,3.698,3.696,\bfseries{3.712}
Detail,3.563,3.583,3.590,\bfseries{3.602}
Language,4.509,4.575,4.618,\bfseries{4.632}
Csense.,0.957,0.973,0.974,\bfseries{0.982}
General,3.517,3.546,3.557,\bfseries{3.572}
\%Sat.,57.8\%,58.5\%,59.2\%,\bfseries{59.3\%}
\end{filecontents*}
\pgfplotstabletypeset[
    col sep=comma,
    string type, 
    every head row/.style={%
        before row=\Xhline{1pt},
        after row=\Xhline{1pt}
    },  
    columns/metric/.style={column name={}, column type={@{\hskip3pt}p{1.1cm}|}},
    columns/validation/.style={column name=No fix, column
    type={@{\hskip3pt}S@{\hskip3pt}|}}, 
    columns/C2_1/.style={column name=Commonsense, column
    type={@{\hskip3pt}C@{\hskip3pt}|}}, 
    columns/C2_2/.style={column name=Language, column
    type={@{\hskip3pt}M|}}, 
    columns/C2_12/.style={column name=All fixes,
    column type={@{\hskip3pt}M}},  
    every last row/.style={after
    row=\Xhline{1pt}} ]{data/c2_fix_human.csv}
\caption{Language Model fixes --- \eval\ dataset.}
\label{tb:fixc2}
\end{table}
\noindent{\bfseries{Language Model fixes.}} 
Language fixes are generally more effective than the
commonsense fixes as they have a higher coverage and they generalize better to other images. Fixes in the Language Model increase
the number of satisfactory captions by only 3\%.
 
\noindent\fbox{\begin{minipage}{0.97\columnwidth} {\bf Result:} The impact of Language Model fixes is limited due to the fact that most
captions with language mistakes also have other problems which cannot be fixed only through this component.
\end{minipage}}
\begin{table}[t]
\centering
\footnotesize
\newcolumntype{C}{>{\centering\arraybackslash}m{3.0cm}}
\newcolumntype{S}{>{\centering\arraybackslash}m{1.1cm}}
\newcolumntype{M}{>{\centering\arraybackslash}m{1.0cm}}
\begin{filecontents*}{data/c3_fix_human.csv}
metric,validation,C3_1
Accuracy,3.674,\bfseries{4.145}
Detail,3.563,\bfseries{3.966}
Language,4.509,\bfseries{4.626}
Csense.,0.957,\bfseries{0.988}
General,3.517,\bfseries{3.973}
\%Sat.,57.8\%,\bfseries{73.6\%}
\end{filecontents*}
\pgfplotstabletypeset[
    col sep=comma,
    string type, 
    every head row/.style={%
        before row=\Xhline{1pt},
        after row=\Xhline{1pt}
    },  
    columns/metric/.style={column name={}, column type={@{\hskip3pt}p{1.5cm}|}},
    columns/validation/.style={column name=No fix, column
    type={@{\hskip3pt}S@{\hskip3pt}|}}, 
    columns/C3_1/.style={column name=Reranking (All fixes), column
    type={@{\hskip3pt}C@{\hskip3pt}}}, 
    every last row/.style={after 
    row=\Xhline{1pt}} ]{data/c3_fix_human.csv}
\caption{Caption Reranker fixes --- \eval\ dataset.}
\end{table}
\noindent{\bfseries{Caption Reranker fixes.}} 
As a final component, fixes in the Caption Reranker directly affect the final caption. This means that if there is a
plausible caption in the Top10 set better than the original best caption, that caption is going to be ranked higher
after the fix and will directly improve the system output.
This explains why Caption Reranker improvements are higher than all other
component fixes.

\noindent\fbox{\begin{minipage}{0.97\columnwidth} {\bf Result:} The system
improves by a factor of 27\% after the Reranker fixes. Although this provides the most effective system
improvement, its influence is limited to instances with at least one satisfactory caption in Top10, which is the case
only for 80.8\% of our dataset.
\end{minipage}}
\begin{table}[t]
\centering
\footnotesize
\newcolumntype{C}{>{\centering\arraybackslash}m{3.0cm}}
\newcolumntype{S}{>{\centering\arraybackslash}m{0.9cm}}
\newcolumntype{M}{>{\centering\arraybackslash}m{1.1cm}}

\begin{filecontents*}{data/c123_fix_human.csv}
metric,validation,C1_1234,C2_12,C3_1,all
Accuracy,3.674,4.035,3.712,4.145,\bfseries{4.451}
Detail,3.563,3.916,3.602,3.966,\bfseries{4.247}
Language,4.509,4.432,4.632,4.626,\bfseries{4.660}
Csense.,0.957,0.942,0.982,0.988,\bfseries{0.998}
General,3.517,3.831,3.572,3.973,\bfseries{4.264}
\%Sat.,57.8\%,68.0\%,59.3\%,73.6\%,\bfseries{86.9\%}
\end{filecontents*}
\pgfplotstabletypeset[
    col sep=comma,
    string type, 
    every head row/.style={%
        before row=\Xhline{1pt},
        after row=\Xhline{1pt}
    },  
    columns/metric/.style={column name={}, column type={@{\hskip3pt}p{1.1cm}|}},
    columns/validation/.style={column name=No fix, column
    type={@{\hskip3pt}S@{\hskip3pt}|}},
    columns/C1_1234/.style={column name=Visual Detector, column
     type={M|}},
    columns/C2_12/.style={column name=Language Model, column
    type={M|}},
    columns/C3_1/.style={column name=Caption Reranker, column
    type={M|}},
    columns/all/.style={column name=All fixes, column
    type={S}}, 
    every last row/.style={after  
    row=\Xhline{1pt}} ]{data/c123_fix_human.csv}
\caption{Complete fix workflow --- \eval\ dataset}
\label{tb:complete_fix_workflow}
\end{table}  

\begin{figure}[h!]
\footnotesize
\begin{tikzpicture}
        \begin{axis}[
          ybar=.1cm,
          bar width = 0.15 cm,
          ymin=0,
          ymax=5.9,
          tickpos=left,
          width=0.57\columnwidth,
    	  ylabel shift = -0.2cm,
    	  xlabel shift = -0.9cm,    	  
    	  height=.5\columnwidth,
          xticklabels={No fix, Visual Detector, Language Model, Caption
          Reranker, All Fixes},
          xticklabel style={rotate=90, text width = 1cm, draw = gray},          
          ylabel=General evaluation,
          tick label style={font=\scriptsize},          
          enlarge x limits=0.1,
          nodes near coords,
          nodes near coords align={vertical},
          every node near coord/.append style={font=\scriptsize, rotate=90,
          anchor=west}, xtick=data,
          name=plot1
         ]            
\addplot[fill=black!25] coordinates {(1, 4.306) (2, 4.171)  (3, 4.267)  (4,
4.424)  (5, 4.48)}; 
\addplot[fill=black!55] coordinates {(1, 2.437) (2, 3.364) (3, 2.619) (4,
3.356) (5, 3.968)};
\end{axis}

\begin{axis}[
  ybar=.1cm,
  bar width = 0.15 cm,
  ymin=0,
  ymax=130,
  tickpos=left,
  width=0.57\columnwidth,
  ylabel shift = -0.2cm,
  xlabel shift = -0.9cm,    	  
  height=.5\columnwidth,
  xticklabels={No fix, Visual Detector, Language Model, Caption
  Reranker, All Fixes},
  xticklabel style={rotate=90, text width = 1cm, draw = gray},          
  ylabel=\% Satisfactory,
  tick label style={font=\scriptsize},          
  enlarge x limits=0.1,
  nodes near coords,
  nodes near coords align={vertical},
  every node near coord/.append style={font=\scriptsize, rotate=90,
  anchor=west}, xtick=data,
  at=(plot1.right of south east), anchor=left of south west
 ]            
\addplot[fill=black!25] coordinates {(1, 100) (2, 83.3)  (3, 95.8)  (4,
 94.5)  (5, 95.8)}; \label{plot:sat}
\addplot[fill=black!55] coordinates {(1, 0) (2, 47.2) (3, 9.2) (4,
45.0) (5, 74.6)}; \label{plot:unsat}
\end{axis}
\matrix[
	matrix of nodes,
    anchor=south,
    draw = gray,
    inner sep=0.2em,
    font=\scriptsize,
    column 1/.style={anchor=base west},
    column 2/.style={anchor=base west}
  ]at(4.0cm, 2.7cm) 
  {
    \ref{plot:sat} 	& \sat 		&[5pt]
    \ref{plot:unsat}	& \unsat 	&[5pt]\\};
\end{tikzpicture}
\caption{Human evaluation scores on the \sat\ and
\unsat\ datasets.}
\label{fig:sat_unsat}
\end{figure}
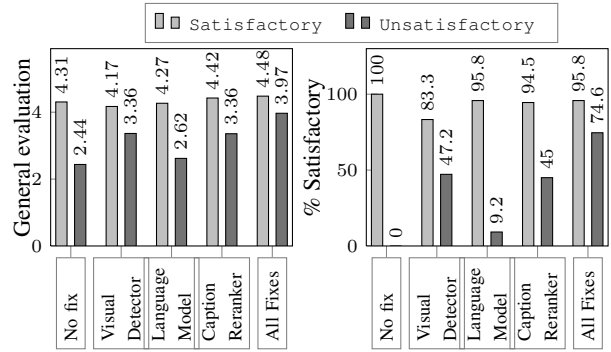
\noindent{\bfseries{Complete fix workflow.}}
Table~\ref{tb:complete_fix_workflow} shows the improvements from each component and the complete fix workflow which
sequentially applies all component fixes. Figure~\ref{fig:sat_unsat} decouples the results for the \sat\ and \unsat\
partitions of the data set. 

\noindent\fbox{\begin{minipage}{0.97\columnwidth} {\bf Result:} The complete fix workflow increases the number of satisfactory captions
by 50\%. In contrast to the initial assumptions of
system designers, fixes in the Caption Reranker are most effective due to the
entanglement in the previous components. Most improvements come from the \unsat\
partition of the dataset. However, because of non-monotonic error behavior in
specific instances, some of the fixes result in slight deteriorations on the \sat\
partition (\emph{e.g.} Visual Detector fixes).
\end{minipage}}

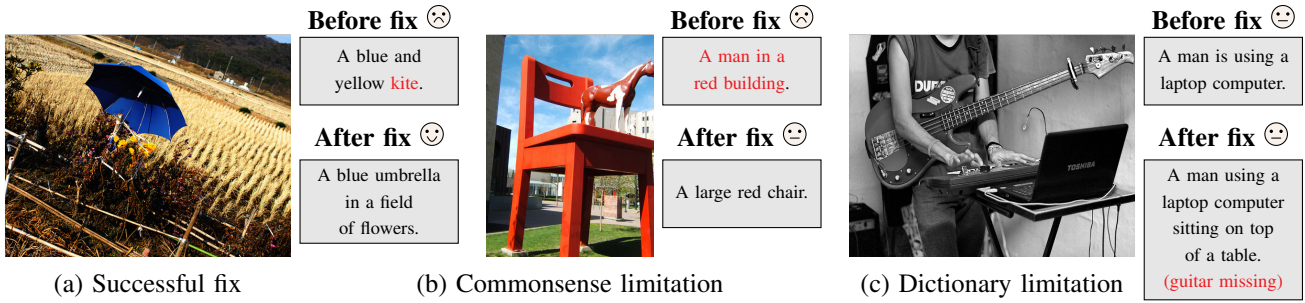
\begin{figure*}[t]
\centering
\begin{tikzpicture}
    \tikzstyle{text}=[draw=none,fill=none]
	\definecolor{light-gray}{gray}{0.9}
	\tikzstyle{boxcaption}=[rectangle,fill=light-gray,minimum width =
	.25\columnwidth, text width = .22\columnwidth, align=center, minimum height = 9 mm, font=\scriptsize, 
	draw=black]
    \node[inner sep=0pt] (img1)  [draw=gray] 
    	{\includegraphics[width=.45\columnwidth,height=.35\columnwidth]{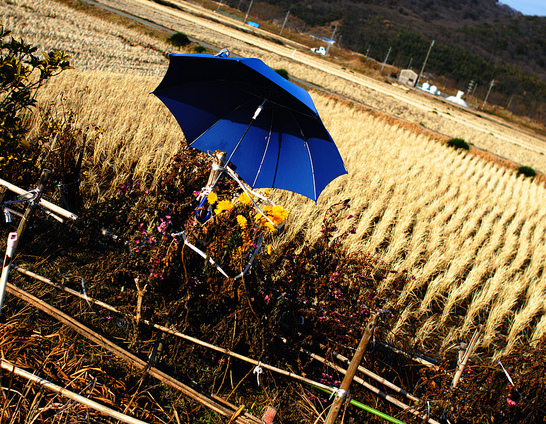}};
   \node[text] (before) [above right=-0.05 cm and 0.05 cm of img1, align=center,
    text width = 20 mm]{\bfseries{Before fix} \Simley{-1}};
   \node[boxcaption] (caption_before) [below=-0.02 cm of before]{A blue and yellow \textcolor{Red}{kite}.};     
   \node[text] (after) [below =0.1 cm of caption_before, align=center,
    text width = 20 mm]{\bfseries{After fix} \Simley{1}};          
    \node[boxcaption] (caption_after) [below=0.05 cm of after]{A blue umbrella in a field of flowers.};
    \node[text] (a) [below =0.1 cm of img1, align=center,
    text width = 40 mm]{(a) Successful fix};          
    
    \node[inner sep=0pt] (img2)  [right=2.6cm of img1, draw=gray] 
    	{\includegraphics[height=.35\columnwidth]{images/examples/COCO_val2014_000000269932.jpg}};
   \node[text] (before2) [above right=-0.05 cm and 0.05 cm of img2, align=center,
    text width = 20 mm]{\bfseries{Before fix} \Simley{-1}};
   \node[boxcaption] (caption_before2) [below=-0.02 cm of before2]{\textcolor{Red}{A man in a red building}.}; 
   
   \node[text] (after2) [below =0.1 cm of caption_before2, align=center,
    text width = 20 mm]{\bfseries{After fix} \Simley{0}};          
    \node[boxcaption] (caption_after2) [below=0.05 cm of after2]{A large red chair.};
   \node[text] (b) [below =0.1 cm of img2, align=center,
    text width = 60 mm]{(b) Commonsense limitation};    
    \node[inner sep=0pt] (img3)  [right=2.6cm of img2, draw=gray] 
    	{\includegraphics[width=.45\columnwidth, height=.35\columnwidth]{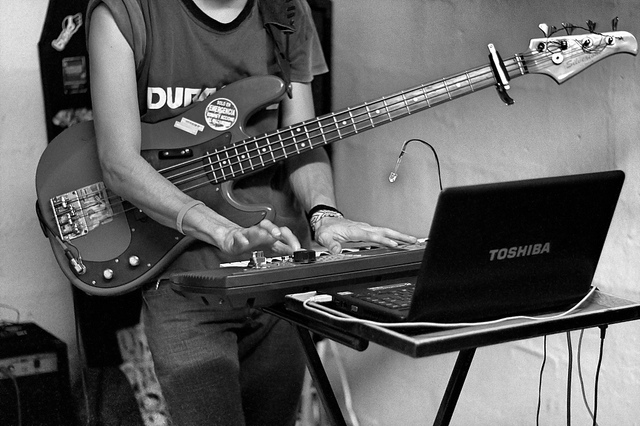}};
   \node[text] (before3) [above right=-0.05 cm and 0.05 cm of img3, align=center,
    text width = 20 mm]{\bfseries{Before fix} \Simley{0}};
   \node[boxcaption] (caption_before3) [below=-0.02 cm of before3]{A man is using a laptop computer.}; 
   
   \node[text] (after3) [below =0.1 cm of caption_before3, align=center,
    text width = 20 mm]{\bfseries{After fix} \Simley{0}};          
    \node[boxcaption] (caption_after3) [below=0.05 cm of after3]{A man using a laptop computer sitting
    on top of a table. \textcolor{Red}{(guitar missing)}};
   \node[text] (b) [below =0.1 cm of img3, align=center,
    text width = 40 mm]{(c) Dictionary limitation};   
\end{tikzpicture}
\caption{Examples of applying the complete fix workflow}
\label{fig:fix_examples}
\end{figure*} 

\subsection{Examples of Fix Integration}
Figure~\ref{fig:fix_examples} presents examples of different ways fix workflows affect
the system output. Figure~\ref{fig:fix_examples}(a) is an example of fixes to the Visual Detector resulting in a
satisfactory system output. In this example, workers removed the erroneous object \texttt{kite} and added
\texttt{umbrella} which propagated the improvement to the final caption. In the larger dataset, successful propagations
of individual component fixes to the final output are also observed for activity fixes, commonsense fixes, and caption
refinement fixes. Figure~\ref{fig:fix_examples}(b) shows an example of fixes having a limited improvement on the final
caption due to the commonsense barrier of the Language Model. In this example, the word \texttt{horse} was present in
both the original and the fixed word list.
However, none of the sentences generated by the Language Model could depict the situation in the image as it was not
found to be likely. This example is not unique, the \unsat\ dataset contains a few more images of the same nature which
describe an unlikely situation that are (at the moment) hard to be described by a statistical automated system.
Figure~\ref{fig:fix_examples}(c) is an example in which improvements from fixes are hindered by the limited size of
the dictionary. Since the word \texttt{guitar} is not in the dictionary, the final caption fails to provide a
satisfactory description.

\section{Discussion} 
\noindent{\bfseries Quality control and methodology cost.}
For all crowdsourcing experiments we applied the following quality control
techniques: (i) spam detection based on worker disagreement, (ii) worker
training via representative examples and detailed instructions, (iii) periodical
batching to prevent worker overload and keep them engaged. These techniques
ensured high-quality data and enabled us to rely on the majority vote
aggregation. Depending on the human computation task, specialized
label aggregation techniques can also be leveraged to further improve data quality.

The cost of human-in-the
loop troubleshooting depends on the number of components, the number of
fixes, the fix workload, and the size of
the dataset to be investigated. Our analysis covered various fix workflows on
all components in the 1000 images \eval\ dataset which showed to be
a good representative of the \val\ dataset. The total
cost of the complete fix workflow (the most expensive one) was \$1,850,
respectively spent in system evaluation (\$250), Visual Detector fixes (\$450), Language Model fixes (\$900), and Caption Reranker fixes (\$250). 
For a more specialized
troubleshooting, the system designer can guide the process towards
components that are prime candidates for improvement or on errors to which
users are most sensitive. We provide further details on quality control
and cost aspects in the appendix.

Our analysis shows that even with a reasonably small subset of data, our
human-in-the loop methodology can efficiently characterize system failure and
identify potential component fixes. Alternative improvement methods
(\emph{i.e.} retraining) require a larger amount of data and
oftentimes do not ensure significant improvements. This can happen due inherent learning barriers in the system or
slow learning curves of underlying algorithms.

\noindent{\bfseries System improvement.}
The results from the methodology provide guidance on next steps to improve the
captioning system. First, improving the Reranker emerges as the most promising direction to pursue.
Second, due to entanglement issues, improvements on the Visual Detector are suppressed by the
shortcomings of the Language Model. Therefore, Visual Detector fixes need to be accompanied with a more capable and
commonsense Language Model. 
Note these insights
cannot be revealed via other methodologies that do not involve human
computation as it is challenging to
automatically simulate improved components without significant engineering effort.

There are multiple ways how human input collected from simulating component fixes can help
with permanently implementing component fixes. Human fixes on the Visual
Detector reveal that the majority of mistakes are false detections. This issue can be addressed by improving model precision. Moreover, the data collected from language
and commonsense fixes can be used for training better language models.
or for immediately filtering out phrases and sentences flagged by workers. 
 Finally, since a common type of reranking error occurs when the component
chooses sentences with words scored low by the Detector, the Reranker can be improved by increasing the weight of the image-caption similarity
score.

\noindent{\bfseries Generalizability.} The general
methodology we presented can be applied to a broad range of component-based
systems that are designed to be modular and their component input / output is
human-interpretable. Even in systems in which these assumptions do not hold in
the functional component design, a new structure can be discovered by logically
separating components in boundaries where data dependencies are guaranteed and
the exchanged data can be analyzed by humans.

Applying the methodology to a new system requires the designer to customize the methodology by identifying component
fixes, defining system quality measures and designing human computation tasks for evaluation and for simulating
component fixes.
In addition to the captioning system, we conceptually applied our methodology to two other systems: (i) question
answering with knowledge bases and web search \cite{yih2015semantic}, and (ii) an email-based reminder for a personal
assistant.
Our feasibility analysis showed that both applications are compatible with the methodology and highlighted that the most
crucial aspect of customization is providing careful and non-ambigious training to crowdsourcing workers tailored to the
system context.
Given the novelty of the proposed human interventions, the resulting
 crowdsourcing tasks are expected to be different from the typical labeling tasks frequently encountered in today's
 platforms. Such problems are usually
more interesting and engaging to crowdsourcing workers.

\noindent{\bfseries Alternative use cases.} In this work, we discuss the benefits of using a human-in-the-loop
methodology for troubleshooting integrative systems by simulating fixes within the existing individual components of a system. The
methodology can additionally be used for further troubleshooting use cases as follows:
\begin{itemize}
  \item \emph{Component prototyping} --- An interesting application of human interventions is to completely simulate the output of a
  component if building it is currently difficult or too expensive. This would allow system designers to
  make feasibility studies before developing a new component. 
  \item \emph{Architectural fixes} --- Our goal in this work was to
	study single-component fixes that precisely follow the current given
	architecture.
	Besides individual component fixes,  a system designer may be interested to
	explore fixes with architectural modifications (\emph{e.g.} exposing the image to the Language Model), which is useful if
	designers are not bound to the initial architecture. Some examples include merging, dividing, or
  introducing new components. In these cases, the task design for human fixes needs to be adjusted according to the
  new architectural specification and the data exchange flow between components. 
  \item \emph{Imperfect fixes} --- In our evaluation, we analyzed fixes that simulate perfect component states.
  However, building flawless components is not always feasible. Therefore, it is realistic to leverage the methodology
  to test imperfect fixes by only partially incorporating the human-generated corrections.
\end{itemize} 
\section{Related Work}
\noindent{{\bfseries Human input for Machine Learning.}}
The contribution of crowdsourcing to machine learning has been mostly limited to the creation of offline data sets for
learning (\emph{e.g.}, \cite{lin2014microsoft,sheng2008get}), with recent interest in active crowd participation to the
development of machine learning algorithms \cite{cheng2015flock,zou2015crowdsourcing,alloy}.
However, there has been only limited work on understanding and diagnosing errors of such systems.
On debugging a single classifier, researchers have developed techniques for a domain expert to interact with the machine
learning process \cite{chakarov2016debugging,kulesza2010explanatory,patel2010gestalt,attenberg2011beat}.
Our work contributes to this line of literature by studying the diagnosis of component-based systems, rather than
individual predictive components, by leveraging the crowd input.

\noindent{{\bfseries Crowdsourcing for Image Captioning.}}
Crowdsourcing is heavily used for collecting data for object recognition and
image captioning \cite{lin2014microsoft,fang2015captions}.
In terms of evaluating the performance of a component-based system, previous work explored replacing different
components with human input to measure the effect of component imperfections on system performance
\cite{parikh2011finding,parikh2011human,yao2015trainable}.
This approach provides information on the current system but does not
offer guidance on system improvements. Our work differs from these studies
in that it evaluates the effect of different component fixes on system
performance to guide decisions on future improvements.

\noindent{{\bfseries Troubleshooting and Blame Assignment.}}
The problems of error diagnosis and blame assignment have been studied for systems whose components
are not machine learned and the state of components is binary through rule-based and model-based diagnosis approaches
\cite{darwiche2000model}.
Breese and Heckerman developed Bayesian networks for predicting the state of each component and for making decisions
about the next repair action to take \cite{breese1996decision}. In recent work,
Sculley et. al., overviewed the challenges of maintaining and improving
real-world machine learning systems highlighting error diagnosis as a critical
task in particular for component-based systems \cite{sculley2015hidden}. 
An alternative way of improving machine learning algorithms is active learning \cite{settles2010active}. Current
techniques are applicable to single components (\emph{e.g.} classifiers) but not to integrative systems with multiple components
which is the focus of this work.


\section*{Conclusion and Future Work }
We reviewed our efforts for troubleshooting component-based machine learning systems. The proposed methodology highlights
the benefits of deeper integration of crowd input on troubleshooting and improving these integrative systems.
Future work directions include the exploration of models that learn from the log data of our methodology to predict
which fixes are most likely to improve the system quality for a given input. Such models can enable algorithms to
query human fixes during system execution for improved system output. Finally, there is an opportunity to develop
generalizable pipelines for automating human-in-the-loop troubleshooting of machine learning systems with reusable
crowdsourcing task templates. Such a pipeline would provide valuable insights on system development and create a
feedback loop in support of continuous improvement.

\subsection{Acknowledgements} This research was initiated during an
internship of Besmira Nushi at Microsoft Research. The work was supported in
part by the Swiss National Science Foundation. The authors would like to thank Xiadong He, Margaret Mitchell, and Larry Zitcnik for their 
support in understanding and modifying the execution of the image captioning
system, as well as Paul Koch for his help while building the necessary
project infrastructure. Our thanks extend to Dan Bohus for the valuable insights
in current problems in integrative systems and Jaime Teevan for helpful discussions on this work.

\clearpage
\bibliographystyle{aaai}
\bibliography{library}
\clearpage
\begin{appendix}

\section{A. Experimental evaluation: automatic scores}
The automatic scores for evaluating image captioning systems are adapted from
automatic machine translation scores where the automatically translated text is
compared to human-generated translations. Similarly, in image captioning, an
automatic caption is compared to the five image captions retrieved from
crowdsourcing workers available in the MSCOCO dataset. While this evaluation is
generally cheaper than directly asking people to report their satisfaction, it
does not always correlate well with human satisfaction. More specifically,
studies show that often what people like is not necessarily similar to what
people generate \cite{vedantam2015cider}. This phenomena happens due to the
fact that an image description can be expressed in many different ways from
different people. However, designing good automatic scores for image
captioning (and machine learning tasks in general) is an active research area
\cite{yang2011corpus,vedantam2015cider,banerjee2005meteor,papineni2002bleu}, as
it facilitates systematic and large-scale evaluation of systems. 

In our evaluation, we experimented with CIDEr, Bleu4,
Bleu1, ROUGEL, and METEOR. The Bleu scores \cite{papineni2002bleu} compute
n-gram precision similarities of the candidate caption with respect to the reference captions. For example, Bleu4
scores higher those captions that have patterns of 4 consecutive words in common
with the human captions. However, Bleu scores do not explicitly model recall
which in this case would measure how well does the automatic caption cover the
content of the human captions. ROUGE \cite{lin2004rouge} instead, is a
recall-based score that was proposed for content summarization which makes it suitable for image captioning
as the caption is in fact a textual summary of the visual content in the image.
METEOR \cite{banerjee2005meteor} and CIDEr \cite{vedantam2015cider} take into
account multiple versions of captions independently and have shown to correlate
better with human satisfaction than the other methods.
\begin{table}[t]
\centering
\footnotesize
\begin{filecontents*}{data/automatic_current_state.csv}
metric,validation_dataset,evaluation_dataset,satisfactory,unsatisfactory
CIDEr,0.909,0.974,1.248,0.628
Bleu 4,0.293,0.294,0.368,0.184
Bleu 1,0.697,0.693,0.757,0.606
ROUGE L,0.519,0.521,0.578,0.443
METEOR,0.247,0.248,0.284,0.2
\end{filecontents*}
\pgfplotstabletypeset[
    col sep=comma,
    string type, 
    every head row/.style={%
        before row={\Xhline{1pt}
            \multicolumn{5}{l}{Automatic Scores}  \\ \hline
        },
        after row=\Xhline{1pt}
    },
    columns/metric/.style={column name=, column type={p{2.7cm}|}},
    columns/validation_dataset/.style={column name=\valshort, column type={c|}},
    columns/evaluation_dataset/.style={column name=\evalshort, column type={c|}},    
    columns/satisfactory/.style={column name=\satshort, column type={c|}},
    columns/unsatisfactory/.style={column name=\unsatshort, column type={c}},    
    every last row/.style={after row=\Xhline{1pt}}
    ]{data/automatic_current_state.csv} 
\caption{Current system evaluation --- Automatic scores.}
\label{tb:currentautomatic}
\end{table}
\begin{table}[t]
\centering
\footnotesize
\newcolumntype{C}{>{\centering\arraybackslash}m{0.9cm}}
\newcolumntype{D}{>{\centering\arraybackslash}m{1.0cm}}
\begin{filecontents*}{data/c1_fix_automatic.csv}
metric,validation,C1_12,C1_34,C1_13,C1_24,C1_1234
CIDEr,0.974,\bfseries{1.048},0.948,0.995,1.023,1.045
Bleu4,0.294,\bfseries{0.298},0.289,0.299,0.289,0.295
Bleu1,0.693,0.713,0.690,0.698,0.711,\bfseries{0.719}
ROUGEL,0.521,\bfseries{0.529},0.517,0.524,0.524,0.528
METEOR,0.248,0.254,0.247,0.251,0.253,\bfseries{0.257}
\end{filecontents*}
\pgfplotstabletypeset[
    col sep=comma,
    string type, 
    every head row/.style={%
        before row={\Xhline{1pt}
            \multicolumn{7}{l}{Automatic Scores --- \eval\ dataset}  \\
            \hline },
        after row=\Xhline{1pt}
    }, 
    columns/metric/.style={column name={}, column type={@{\hskip3pt}p{1.1cm}|}},
    columns/validation/.style={column name=No fix, column
    type={@{\hskip3pt}C@{\hskip3pt}|}}, 
    columns/C1_12/.style={column name=Object, column
    type={@{\hskip3pt}C@{\hskip3pt}|}}, columns/C1_34/.style={column name=Activity, column
    type={@{\hskip3pt}C|}}, columns/C1_13/.style={column name=Addition,
    column type={@{\hskip3pt}D|}},  
    columns/C1_24/.style={column name=Removal, column
    type={@{\hskip3pt}D|}}, columns/C1_1234/.style={column name=All fixes,
    column type={@{\hskip3pt}C@{\hskip3pt}}}, every last row/.style={after
    row=\Xhline{1pt}} ]{data/c1_fix_automatic.csv}
\caption{Visual Detector fixes --- \eval\ dataset.}
\label{tb:fixc1automatic} 
\end{table} 
\begin{table}
\centering
\footnotesize
\newcolumntype{C}{>{\centering\arraybackslash}m{1.9cm}}
\newcolumntype{S}{>{\centering\arraybackslash}m{0.9cm}}
\newcolumntype{M}{>{\centering\arraybackslash}m{1.5cm}}
\begin{filecontents*}{data/c2_fix_automatic.csv}
metric,validation,C2_1,C2_2,C2_12
CIDEr,0.974,0.975,\bfseries{0.983},0.98
Bleu4,0.294,0.297,0.297,\bfseries{0.298}
Bleu1,0.693,\bfseries{0.694},0.690,0.691
ROUGEL,0.521,0.524,0\bfseries{.526},0.527
METEOR,0.248,\bfseries{0.249},0.247,0.247
\end{filecontents*}
\pgfplotstabletypeset[
    col sep=comma,
    string type,
    every head row/.style={%
        before row={\Xhline{1pt}
            \multicolumn{5}{l}{Automatic Scores --- \eval\ dataset}  \\
            \hline },
        after row=\Xhline{1pt}
    },
    columns/metric/.style={column name={}, column type={@{\hskip3pt}p{1.1cm}|}},
    columns/validation/.style={column name=No fix, column
    type={@{\hskip3pt}S@{\hskip3pt}|}},
    columns/C2_1/.style={column name=Commonsense, column
    type={@{\hskip3pt}C@{\hskip3pt}|}},
    columns/C2_2/.style={column name=Language, column
    type={@{\hskip3pt}M|}},
    columns/C2_12/.style={column name=All fixes,
    column type={@{\hskip3pt}M}},
    every last row/.style={after
    row=\Xhline{1pt}} ]{data/c2_fix_automatic.csv} 
\caption{Language Model fixes --- \eval\ dataset.}
\label{tb:fixc2automatic}
\end{table}

\begin{table}
\centering
\footnotesize
\newcolumntype{C}{>{\centering\arraybackslash}m{3.0cm}}
\newcolumntype{S}{>{\centering\arraybackslash}m{1.1cm}}
\newcolumntype{M}{>{\centering\arraybackslash}m{1.0cm}}
\begin{filecontents*}{data/c3_fix_automatic.csv}
metric,validation,C3_1
CIDEr,0.974,\bfseries{1.087}
Bleu4,0.294,\bfseries{0.320}
Bleu1,0.693,\bfseries{0.720}
ROUGEL,0.521,\bfseries{0.543}
METEOR,0.248,\bfseries{0.261}
\end{filecontents*}
\pgfplotstabletypeset[
    col sep=comma,
    string type, 
    every head row/.style={%
        before row={\Xhline{1pt}
            \multicolumn{3}{l}{Automatic Scores --- \eval\ dataset}  \\ 
            \hline },
        after row=\Xhline{1pt}
    },  
    columns/metric/.style={column name={}, column type={@{\hskip3pt}p{1.5cm}|}},
    columns/validation/.style={column name=No fix, column
    type={@{\hskip3pt}S@{\hskip3pt}|}}, 
    columns/C3_1/.style={column name=Reranking (All fixes), column
    type={@{\hskip3pt}C@{\hskip3pt}}}, 
    every last row/.style={after 
    row=\Xhline{1pt}} ]{data/c3_fix_automatic.csv}
\caption{Caption Reranker fixes --- \eval\ dataset.} 
\label{tb:fixc3automatic}
\end{table}

\begin{table}[t!]
\centering
\footnotesize
\newcolumntype{C}{>{\centering\arraybackslash}m{3.0cm}}
\newcolumntype{S}{>{\centering\arraybackslash}m{0.9cm}}
\newcolumntype{M}{>{\centering\arraybackslash}m{1.1cm}}
\begin{filecontents*}{data/c123_fix_automatic.csv}
metric,validation,C1_1234,C2_12,C3_1,all
CIDEr,0.974,1.045,0.98,1.087,\bfseries{1.106}
Bleu4,0.294,0.295,0.298,0.320,\bfseries{0.315}
Bleu1,0.693,0.719,0.691,0.720,\bfseries{0.743}
ROUGEL,0.521,0.528,0.527,0.543,\bfseries{0.543}
METEOR,0.248,0.257,0.247,0.261,\bfseries{0.266}
\end{filecontents*}
\pgfplotstabletypeset[
    col sep=comma,
    string type, 
    every head row/.style={%
        before row={\Xhline{1pt}
            \multicolumn{6}{l}{Automatic Scores --- \eval\ dataset}  \\
            \hline },
        after row=\Xhline{1pt}
    },  
    columns/metric/.style={column name={}, column type={@{\hskip3pt}p{1.1cm}|}},
    columns/validation/.style={column name=No fix, column
    type={@{\hskip3pt}S@{\hskip3pt}|}},
    columns/C1_1234/.style={column name=Visual Detector, column
     type={M|}},
    columns/C2_12/.style={column name=Language Model, column
    type={M|}},
    columns/C3_1/.style={column name=Caption Reranker, column
    type={M|}},
    columns/all/.style={column name=All fixes, column
    type={S}}, 
    every last row/.style={after  
    row=\Xhline{1pt}} ]{data/c123_fix_automatic.csv}
\caption{Complete fix workflow --- \eval\ dataset}
\label{tb:complete_fix_workflow_automatic}
\end{table}  
\noindent {\bfseries The current system state.}
(Table~\ref{tb:currentautomatic}) 
Since MSCOCO contains five human generated captions for all
images in the \val\ dataset (40,504 images), we could compute the automatic
scores for the whole \val\ dataset and compare it with the scores from \eval. 

\noindent\fbox{\begin{minipage}{0.97\columnwidth} {\bf Result:} The
evaluation on automatic scores shows that the \eval\ dataset is a good
representative of the whole \val\ dataset as it has similar automatic scores.
The highest differences between the \sat\ and \unsat\ datasets are observed for
the CIDEr and Bleu4 scores, highlighting their agreement with human
satisfaction.
\end{minipage}} 

\noindent{\bfseries{Component fixes.}}
(Tables~\ref{tb:fixc1automatic}, \ref{tb:fixc2automatic},
\ref{tb:fixc3automatic}, \ref{tb:complete_fix_workflow_automatic}) 
Among all automatic scores, CIDEr, Bleu4, and METEOR preserve the same trends as the human satisfaction score. For
example, they confirm that object and removal fixes are more effective than
respectively the activity and addition fixes. The non-monotonic error
propagations are also reflected in the automatic score analysis as
previously concluded from the human satisfaction evaluation. In overall,
Caption Reranker fixes remain the most effective ones, followed by the Visual
Detector, and the Language Model.

Improvements for the ROUGEL, METEOR, and Bleu4 are lower as the metrics require a more complete coverage of the
human captions which is challenging to achieve in a sentence with limited
length. For example, ROUGEL and METEOR require high recall, while
Bleu4 relies on exact matches of 4-gram patterns.
\begin{table*}[t]
\centering
\footnotesize
\begin{tabular}{l r r r r}
  \Xhline{1pt}
  Crowdsourcing task	&No. tasks per workflow	&Cost / assignment	&No.
  assignments	&Total max. cost \\ \Xhline{1pt}
  
  Visual Detector --- Add / remove objects	&1000
  &\$0.05	&3	&\$250.00 \\
  
  Visual Detector --- Add / remove activities	&1000	&\$0.04	&3
  &\$200.00 \\ \hline
  
  Language Model --- Mark noncommonsense captions	&$\leq$ 10000	&\$0.02	&3
  &\$600.00 \\
  
  Language Model --- Remove non-fluent captions	& $\leq$ 10000	&\$0.01	&3
  &\$300.00 \\ \hline
  
  Caption Reranker --- Rerank Top 10 captions	&1000	&\$0.05	&5	&\$250.00 \\
  \hline
  
  System evaluation --- Human satisfaction	&1000	&\$0.05	&5	&\$250.00 \\
  \Xhline{1pt}
  
  Maximum workflow cost &	& 	& 	&\$1,850.00 \\
					
   \Xhline{1pt}
    \end{tabular}
    \caption{Summary of the crowdsourcing cost in the \eval\ dataset.}
    \label{tb:cost}
\end{table*}

\section{B. Quality control and methodology cost}

\noindent{\bfseries{Quality control.}} For all crowdsourcing
experiments we applied the following quality control techniques: 

\noindent \emph{Worker training} --- Providing accurate instructions to
crowdsourcing workers is a crucial aspect in applying our methodology to new machine learning
systems. We paid special attention to the task setup by reiterating over the
user interface design, providing multiple examples of correct solutions, and
giving online feedback to workers on their work quality. Detailed task
instructions are necessary for workers to understand the goal of the system and
the role of their fixes. We also noticed that workers get more engaged if they
understand how their work contributes to improving the presented application. 

\noindent \emph{Spam detection} --- Due to the multimodal nature of the case
study, the set of crowdsourcing tasks we used in our methodology is rich and versatile
both in terms of content  (\emph{e.g.} images, words, sentences) and design
(\emph{e.g.} yes / no questions, word list correction etc.). Therefore, each
task required specific spam detection techniques. Given the lack of ground
truth, we used worker disagreement as the main criterion to distinguish
low-quality work. However, due to potential subjectivity in our tasks, we
used worker disagreement only as an initial indication of low quality and
further analyzed distinct cases to decide upon work acceptance.

\noindent \emph{Batching} --- Large crowdsourcing batches are exposed to
low-quality work risk as workers may get tired or reinforce repetitive and wrong biases over
time.
To avoid such effects, we performed experiments in small-sized batches (\emph{e.g.} 250
image-caption pairs) which were published in periodical intervals. This allowed
us to analyze the batch data offline and give constructive feedback to specific
workers on how they can improve their work in the future. The strategy also
helped to keep workers engaged and motivated.

\noindent{\bfseries{Methodology cost.}} Table~\ref{tb:cost} shows the cost
of each crowdsourcing task used in our methodology for the captioning system.
Based on this table, system designers can estimate the cost of any future fix workflow. The
 last column corresponds to the total cost for all
instances in the \eval\ dataset. Note that the data collected for the Language
Model fixes is reusable as the sentences pruned in one workflow can be safely
pruned in other workflows as well. Once we executed the tasks for the workflow
that fixes only the Language Model, we observed that due to self-repetition, the
number of new sentences generated from other workflows is at least 4 times
lower. This means that the cost of fixes for the Language Model continuously
decreases over time.

For a new machine learning sytem, the associated cost to human-in-the
loop troubleshooting depends on the number of components, the number of
fixes per component, the fix workload, and the size of
the dataset to be investigated.
\end{appendix}
\end{document}